\documentclass[conference]{IEEEtran}
\IEEEoverridecommandlockouts
\usepackage{cite}
\usepackage{amsmath,amssymb,amsfonts}
\usepackage{algorithmic}
\usepackage{graphicx}
\usepackage{textcomp}
\usepackage{xcolor}

\usepackage{url}
\usepackage{hyperref}

\usepackage{array}
\usepackage{booktabs} 

\usepackage{multirow}

\def\BibTeX{{\rm B\kern-.05em{\sc i\kern-.025em b}\kern-.08em
    T\kern-.1667em\lower.7ex\hbox{E}\kern-.125emX}}
\begin{document}
\title{TROJAN-GUARD: Hardware Trojans Detection Using GNN  in RTL Designs\\


\author{%
Kiran Thorat\IEEEauthorrefmark{1},
Amit Hasan\IEEEauthorrefmark{1},
Caiwen Ding\IEEEauthorrefmark{2},
Zhijie Shi\IEEEauthorrefmark{1}
\\[0.5ex]
\IEEEauthorblockA{%
\IEEEauthorrefmark{1}Computer Science and Engineering Department, University of Connecticut\\
\{kiran\_gautam.thorat, amit.hasan, zshi\}@uconn.edu}
\IEEEauthorblockA{%
\IEEEauthorrefmark{2}Department of Computer Science \& Engineering, University of Minnesota \\
dingc@umn.edu}
}

 \thanks{This work was supported by CHEST – NSF IUCRC Center for Hardware and Embedded Systems Security and Trust (CNS-1916756 with Project No. P19\_23).}
}

\maketitle

\begin{abstract}
Chip manufacturing is a complex process, and to achieve a faster time to market, an increasing number of untrusted third-party tools and designs from around the world are being utilized. The use of these untrusted third party intellectual properties (IPs) and tools increases the risk of adversaries inserting hardware trojans (HTs). The covert nature of HTs poses significant threats to cyberspace, potentially leading to severe consequences for national security, the economy, and personal privacy. Many graph neural network (GNN)-based HT detection methods have been proposed. However, they perform poorly on larger designs because they rely on training with smaller designs. Additionally, these methods do not explore different GNN models that are well-suited for HT detection or provide efficient training and inference processes. We propose a novel framework that generates graph embeddings for large designs (e.g., RISC-V) and incorporates various GNN models tailored for HT detection. Furthermore, our framework introduces domain-specific techniques for efficient training and inference by implementing model quantization. Model quantization reduces the precision of the weights, lowering the computational requirements, enhancing processing speed without significantly affecting detection accuracy. We evaluate our framework using a custom dataset, and our results demonstrate a precision of 98.66\% and a recall (true positive rate) of 92.30\%, highlighting the effectiveness and efficiency of our approach in detecting hardware trojans in large-scale chip designs.


\end{abstract}


\section{Introduction}

The widespread presence of hardware chips in modern computing systems highlights its critical importance. The hardware chips are now integrated into diverse applications such as smartphones, IoT, AI, and autonomous vehicles \cite{saha2024llm, fu2023llm4sechw,ray2017system}. As these hardware chips collect, analyze, and store the data, security concerns regarding them are increasing. Furthermore, hardware chips are increasingly becoming larger, and more complex which makes hardware chips vulnerable to a plethora of attacks. For example,  hardware trojan \cite{tehranipoor2010survey}, side-channel attacks \cite {he2017secure, pundir2022power}, information leakage \cite{nahiyan2016avfsm, contreras2017security}, and fault injection attacks \cite{nahiyan2018security}. These advanced attacks can threaten the risk of data confidentiality, integrity, and reliability of computing systems.
 The vulnerabilities come from various sources such as malicious insider tampering, unintentional design errors, weak testing and verification frameworks, and optimization by computer-aided design (CAD) tools that overlooked security implications \cite{mishra2017security}.
 Unlike software bugs which can be mitigated after the software deployment, the vulnerabilities in hardware chips are often non-mitigable after the chips are manufactured. In addition, to reduce the cost, it is preferable to detect and mitigate vulnerabilities as early as possible, for example, in the register transfer logic (RTL) stage.

Driven by time-to-market demands, hardware chip producers often outsource designs to third parties and also use computer-aided design (CAD) tools from different vendors. However, the globalization of the CAD tool industry introduces the risk of hardware trojan (HT) insertion by malicious entities. Ensuring trustworthy hardware chip designs thus requires reliable HT detection methods. Graph Neural Networks (GNNs) have attracted increasing interest because of their performance in graph-based learning tasks \cite{kipf2017semisupervisedclassificationgraphconvolutional}. In hardware chip design, netlists can be represented as graphs, making GNNs a promising choice in several chip design tasks including HT detection.

Existing HT detection studies lacks the coverage of different hardware designs in their datasets. Many use the Trusthub \cite{TrustHubTrojan} benchmark, but only select a subset of designs (e.g., AES, PIC, and RS232). For instance, GNN4HT \cite{chen2024gnn4ht} uses two-stage GNN approach but evaluates only on 21 hardware designs, while \cite{10560302} generates additional designs for data balancing yet still evaluates on only AES, PIC, and RS232. Moreover, most existing studies use Graph Convolutional Networks (GCNs) \cite{kipf2017semisupervisedclassificationgraphconvolutional} without considering other GNN variants that might be better suited for HT detection. For example, HW2VEC \cite{yu2021hw2vecgraphlearningtool} relies on a GCN-based model. Additionally, model compression strategies such as sparsity and quantization, which can improve training and inference efficiency, have not been explored in this domain.

To address these gaps, we propose a novel GNN-based framework for HT detection. Our main contributions include:


\begin{itemize}
\item We create a new dataset by injecting multiple hardware trojan variants into a diverse collection of RTL designs, giving broad design coverage.
\item We design GNN architectures tailored to HT detection by systematically comparing GCN, GraphSAGE, GAT, and GIN layers for netlist-based EDA graphs.

 \item We develop a 4‑bit post‑training quantization scheme for GNN‑based HT detectors, trimming model size by $8\times$ and speeding up inference while preserving the close to original detection accuracy.
\end{itemize}


We evaluate proposed framework on the new dataset and achieve a precision of 98.66\,\% and a recall of 92.30\,\%, demonstrating the framework’s ability to detect HTs in large‑scale chip designs.  
The remainder of this paper is organized as follows: Section\,II reviews background and related work; Section\,III explains the trojan‑injection procedure and resulting dataset, details the GNN architecture and the 4‑bit quantization pipeline; Section\,IV presents experimental results and discussion.

\section{Background and Related work}
Hardware trojans are malicious modifications made to hardware chips that can threaten national security, the economy, and personal privacy. Detecting HTs early in the design process, especially at the RTL, is important because it is more cost-effective in the early stages. As hardware chips become more complex and are used in various applications such as smartphones, IoT devices, AI systems, and autonomous vehicles, the need for effective HT detection methods has increased. Traditional ML methods have been widely used for HT detection by analyzing structural and functional features extracted from circuit designs. These methods often rely on manually created features from gate-level netlists or other design representations. For example, some studies have used neural networks (NNs) and random forests (RFs) to classify nodes in gate-level netlists, achieving good true positive rates (TPR) and true negative rates (TNR) \cite{kurihara2020evaluation}. Similarly, other research has used models like eXtreme Gradient Boosting (XGBoost) with selected structural features to improve classification performance \cite{negishi2024hardware}. However, these traditional ML models face challenges such as the need for extensive feature engineering, difficulties in scaling with increasing circuit complexity, and the lack of a standard feature set for HT detection. GNNs offer a promising alternative to traditional ML models by using the inherent graph structure of circuit designs. GNNs can automatically learn and update node feature representations through message passing and aggregation, reducing the need for manual feature extraction \cite{hasegawa2023node}. This ability allows GNNs to generalize better to new circuit designs and adapt to different structural patterns without significant changes. GNNs \cite{kipf2017semisupervisedclassificationgraphconvolutional} can perform tasks such as node classification, link prediction, and graph classification, making them suitable for identifying HTs. Several studies have shown the effectiveness of GNNs in HT detection. For instance, some approaches have used Graph Convolutional Networks (GCNs) and Graph Attention Networks (GATs) on netlists, achieving high accuracy in identifying trojan nodes \cite{chen2024gnn4ht, ma2024gnn, yasaei2022hardware, yasaei2021gnn4tj}. Other research has used centrality measures from social network analysis, like betweenness centrality (BC) and PageRank (PR), to enhance the detection capabilities of GNN models \cite{hashemi2022graph}. These methods show that GNNs can capture both local and global structural properties of circuits, providing a more thorough analysis for HT detection.
Despite the progress with GNN-based methods, there are still limitations in current approaches. Many existing studies use limited datasets that do not represent the full range of hardware designs, often testing models on a small set of benchmark circuits from TrustHub \cite{TrustHubTrojan}. Additionally, most research has focused on specific types of GNN architectures, such as GCNs, without exploring other variants that might offer better performance or efficiency \cite{yasaei2022hardware}. Furthermore, techniques for model compression, including sparsity and quantization, which could improve the scalability and deployment of GNNs in resource-constrained environments, have not been thoroughly explored in the context of HT detection \cite{ma2024gnn, chen2024gnn4ht}.
To address these challenges, we propose a novel framework that generates graph embeddings for large hardware designs, such as RISC-V cores, and incorporates various GNN models tailored for HT detection. Our framework introduces domain-specific techniques for efficient training and inference by implementing model quantization and inducing sparsity. Model quantization reduces the precision of the weights, lowering computational requirements, while sparsity decreases the number of active connections, enhancing processing speed without significantly affecting detection accuracy. By combining these techniques, our framework achieves both high precision and recall in detecting hardware trojans, making it suitable for large-scale and real-time applications.
\begin{figure*}[ht]
    \centering
    \includegraphics[width=0.98\textwidth]{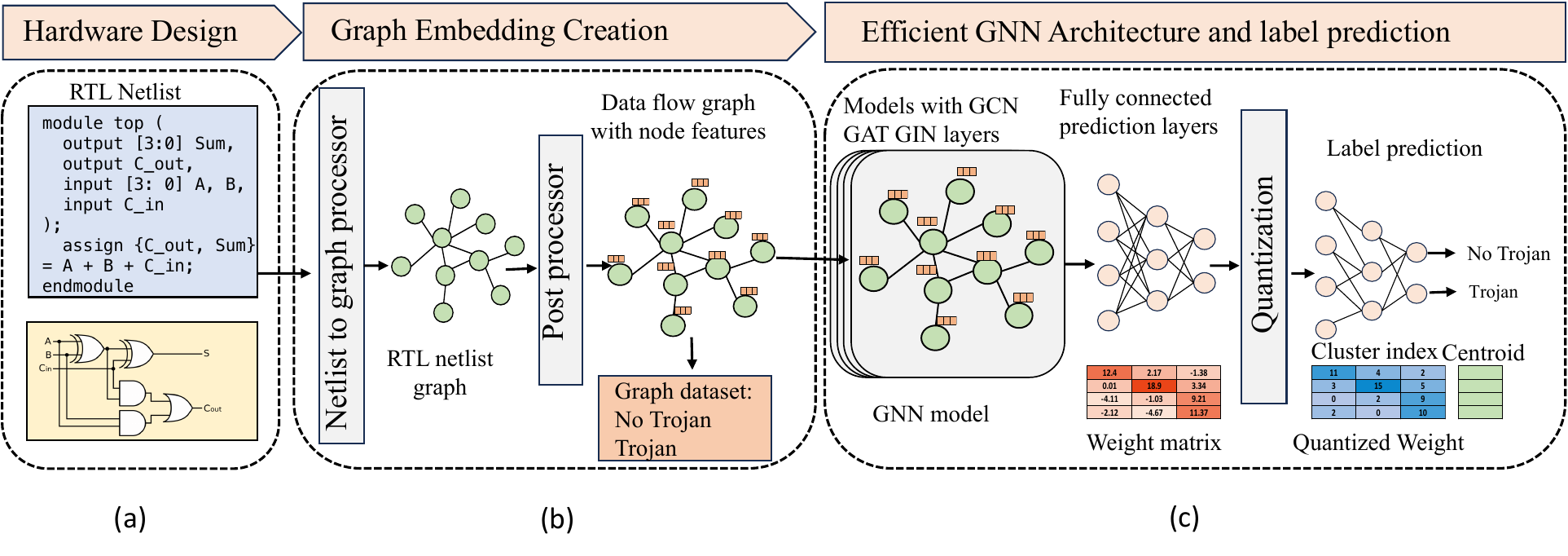}
    \caption{Framework: (a) RTL hardware design (example, full addder, (b) Graph embedding with node features creation, (c) GNN models and efficient training and inference for prediction.}
    \label{fig:overview}
\end{figure*}

\section{Methodology}

\subsection{Hardware Design}
We begin our framework with hardware designs written in RTL Verilog. In chip design, after high-level exploration, the netlist is created using hardware description languages like VHDL or Verilog. We use hardware designs described in RTL Verilog to build the RTL netlist graph. As shown in Fig.~\ref{fig:overview}, we have a full adder example described in an RTL Verilog netlist.

\subsubsection{Hardware Trojan Insertion}
Hardware trojans usually have two parts: a trigger and a payload. Both parts are often inserted into low-activity, small modules of the RTL \cite{chen2024gnn4ht}.

\begin{figure}[htbp!]
\centering
\includegraphics[width=0.99\linewidth]{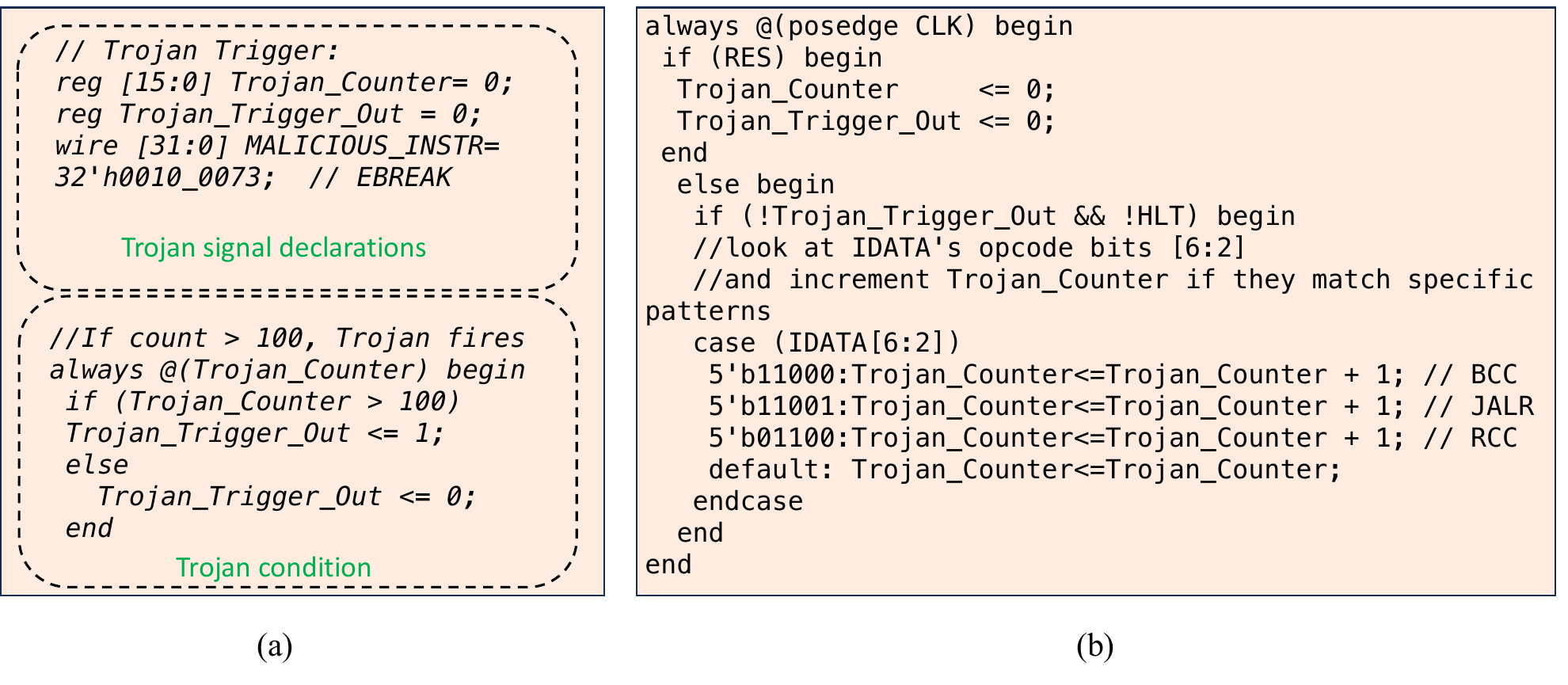}
\caption{(a) Trojan signal declaration and condition, (b) Trojan trigger}
\label{fig:trojan_trigger}
\end{figure}

In the trigger, additional Verilog code triggers signals to activate the payload. As shown in Fig.~\ref{fig:trojan_trigger} (a), we create two registers: \texttt{Trojan\_Counter} (16 bits) and \texttt{Trojan\_Trigger\_Out} in the top module of the RISC-V architecture. If \texttt{Trojan\_Counter} goes above 100, we set \texttt{Trojan\_Trigger\_Out} = 1.

As shown in Fig.~\ref{fig:trojan_trigger} (b), the always block increments \texttt{Trojan\_Counter} when reset is not active and \texttt{IDATA} [6:2] matches certain opcodes (for example, BCC, JALR, RCC).

\begin{figure}[htbp!]
\centering
\includegraphics[width=0.99\linewidth]{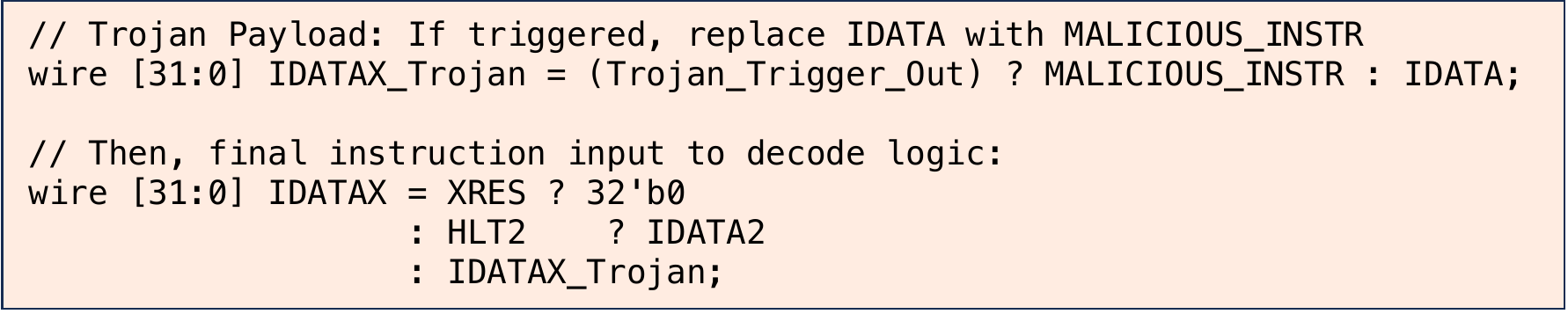}
\caption{Trojan payload}
\label{fig:trojan_payload}
\end{figure}

In the payload, once \texttt{Trojan\_Trigger\_Out} is set, as shown in Fig.~\ref{fig:trojan_payload}, the code overrides \texttt{IDATA} with \texttt{MALICIOUS\_INSTR} = \texttt{32'h0010\_0073}, which corresponds to \texttt{EBREAK}. This disrupts the CPU (like sleep). We mux the real \texttt{IDATA} with the malicious opcode based on \texttt{Trojan\_Trigger\_Out}.
This type of trojans is used for denial of service.
Other types of the trojans we inserted into our diverse set of hardware designs include information leakage, functionality change, and performance degradation.

\subsection{Graph Embedding Creation}
The hardware design is an RTL netlist which is similar to a graph. As shown in Fig. \ref{fig:overview} (b), we use a netlist to graph  processor to create RTL netlist graph, which shows the relationships and dependencies among signals and gives a fundamental expression of the code's computational structure. An RTL design may have multiple modules in separate files, so we need a preprocess phase to flatten the design and fix syntax incompatibilities. Next, we use a toolkit called Pyverilog~\cite{TakamaedaYamazaki2015PyverilogAP}, where a parser extracts an abstract syntax tree from Verilog. This tree goes to the data flow analyzer, which builds a tree for each signal in the circuit, with that signal as the root node. We then merge all these signal DFGs into one graph for the entire circuit, and we trim any disconnected sub-graphs and redundant nodes in the merge phase.

As shown in Fig. \ref{fig:overview} (b), the final DFG based graph is a created with node features. We construct using a directed graph that shows data dependencies from output signals (root nodes) to input signals (leaf nodes). We define it as \(G = (V, E)\), where \(V\) is the set of vertices and \(E\) is the set of directed edges. Let \(V = \{v_1, v_2, \dots, v_n\}\), where each \(v_i\) can be a signal, a constant value, or an operation such as \texttt{xor}, \texttt{and}, concatenation, branch, or branch condition. We define \(E = e_{ij}\) for all \(i, j\) such that \(e_{ij}\in E\) if the value of \(v_i\) depends on the value of \(v_j\), or if the operation \(v_j\) is applied to \(v_i\).



\subsection{Graph Neural Networks}

GNNs are used in many applications across various domains by modeling graph structures \cite{song2024no, kipf2017semisupervisedclassificationgraphconvolutional, kunal2023gnn, NEURIPS2023_12da92b7,zhang2025hydrogen}. They address tasks at the node, edge, or graph level using a consistent two-step procedure: aggregation followed by update.
In the aggregation step, the network collects information from a target node's neighbors.
Given node representations at the $t$-th layer $h^t_u$ and the set of neighbors of target node $v$, $N_v$, the aggregation function at time $t+1$ can be written as:
\begin{equation}
    a^{(t+1)} = \text{AGGREGATE}^{(t+1)}\left( h^{(t)}_u : u \in N_v \right)
\end{equation}

Following aggregation, the update step combines the existing node representation with the aggregated information to update the target node's representation
\begin{equation}
    h^{(t+1)}_v = \text{UPDATE}^{(t+1)}\left(h^{(t)}, a^{(t+1)}\right)
\end{equation}
.

\begin{figure}[htbp!]
\centering
\includegraphics[width=0.95\linewidth]{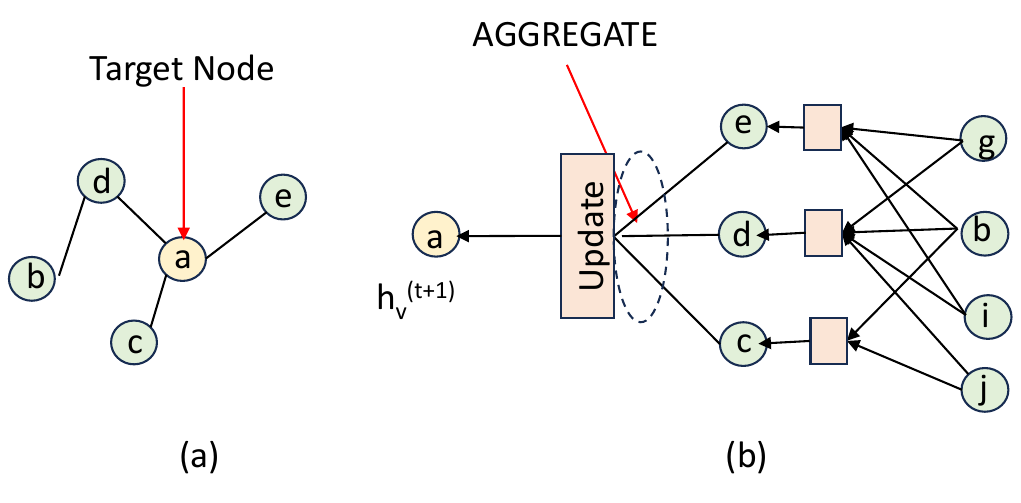}
\caption{(a) Input Graph, (b) GNN: Node aggregate and update}
\label{fig:gnn}
\end{figure}

 Figure \ref{fig:gnn} shows the process of the ``aggregate" and ``update" mechanism in GNNs. Figure \ref{fig:gnn} (a) shows the input graph where the target node $a$ is connected with neighboring nodes $c$, $d$, and $e$. The message-passing happens through the edges that connect the nodes. Figure \ref{fig:gnn} (b) shows the neighboring nodes pass message (node features) to node $a$. GNN does aggregate operation and node features and updates the node features of node $a$.  
 
Different GNN architectures emerge from the choices made for the functions $\text{AGGREGATE}^{(t+1)}(\cdot)$ and $\text{UPDATE}^{(t+1)}(\cdot)$ \cite{vatter2024choosing}. The subsequent sections explore various methods that implement these functions

\subsubsection{Graph Attention Network (GAT)}
GATs \cite{velivckovic2017graph} replace traditional graph convolutions with masked self-attentional layers. These layers assign different weights to each neighbor during the aggregation of neighboring features. This mechanism allows nodes to focus on more relevant neighbors and helps the model capture complex relationships within the graph. Another advantage is that the importance of neighboring nodes is determined dynamically, without requiring prior knowledge of the entire graph structure.

The updated node representations are obtained as follows:
\begin{equation}
h_v^{(t+1)} = \sigma\left(\sum_{k=1}^K \sum_{u \in N_v} \alpha_{vu}^k W_k h_u^{(t)}\right)
\end {equation}
Here, \(\alpha_{vu}^k\) represents the normalized attention coefficient for the \(k\)-th attention head, and \(W_k\) is the corresponding weight matrix. By using attention, the model effectively detects both local and global dependencies.

\subsubsection{Graph Convolutional Network (GCN)}
A commonly used GNN architecture is the Graph Convolutional Network (GCN) \cite{kipf2017semisupervisedclassificationgraphconvolutional, ma2020online}. Inspired by image convolutions, GCNs use convolution filters that operate directly on the graph structure. Unlike images, the neighborhood size of a node in a graph varies. A parameter matrix transforms the node representations obtained from the previous layer, and the transformed representations are weighted according to the graph's adjacency matrix. The update step using GCN is defined as
\begin {equation}
H^{(t+1)} = \sigma\left(\hat{A} H^{(t)} W\right)
\end {equation}
where \(H^{(t+1)}\) denotes the matrix of stacked node representations at layer \(t+1\), \(H^{(t)}\) represents the node representations at layer \(t\), \(\sigma\) is an activation function (e.g., ReLU), \(\hat{A}\) is the normalized adjacency matrix, and \(W\) is the parameter matrix. GCN uses a shared weight for all edges, making the model relatively simple. For more complex graph structures, this approach may offer less expressiveness.

\subsubsection{Graph Isomporhism Network (GIN)}

Another variant is GIN \cite{xu2018powerful}, which uses multi-layer perceptrons (MLPs) to learn the parameters of the update function. It draws inspiration from the Weisfeiler-Lehman (WL) graph isomorphism test \cite{leman1968reduction}, which determines the similarity between two graphs. The node update is defined by
\begin {equation}
h_v^{(t+1)} = \text{MLP}^{(t+1)}\Big((1 + \epsilon^{(t+1)}) \cdot h_v^{(t)} + \sum_{u \in N(v)} h_u^{(t)}\Big)
\end {equation}
where \(\epsilon^{(t+1)}\) is a learnable parameter. GIN uses a simple sum operator to aggregate features, which makes it computationally efficient. The learnable parameter \(\epsilon\) allows the model to adapt to various graph structures and effectively capture graph-level features, making GIN a common choice for graph-level tasks.

\subsection{Efficient training and inference}
Quantization is a commonly used technique to reduce the precision of neural network weights and activations. Thus, the model consumes less memory and with reduced precision, computations become faster. Given a full-precision floating-point weight matrix $W \in \mathbb{R}^{m \times n}$, we create a quantized version $\hat{W}$ by mapping each element to a lower-bit representation:
\begin{equation}
\hat{W} = \text{round}(W / S) \times S,
\end{equation}
where $S$ is a scaling factor. $S$ adjusts the range of the quantized values to match the distribution of the original weights by determining an appropriate step size for rounding, minimizing the error introduced by quantization while ensuring that the values remain representative of the full precision data.

For hardware Trojan detection using GNNs (as shown in Figure~\ref{fig:overview} (a)), quantization makes our workflow more efficient. We apply 4-bit quantization using the \texttt{bitsandbytes} library \cite{dettmers2023spqr}, which reduces our 32-bit model weights to 4-bit precision while keeping accuracy nearly the same. This process reduces the memory footprint, speeds up inference, makes it easier to deploy the model on hardware with limited resources. In 4-bit quantization, weights are restricted to $2^4 = 16$ levels and we can cut storage needs by a factor of 8 compared to 32-bit floating-point weights \cite{Jacob2017QuantizationAT, Dettmers2023QLoRAEF}.

In our work, applying 4-bit quantization helps process and store the model efficiently while keeping accuracy high. Given a graph representation with node features $X$ and an adjacency matrix $A$, a graph convolution operation is performed as follows:
\begin{equation}
H^{(l+1)} = \sigma( A H^{(l)} W^{(l)} ),
\end{equation}
where $W^{(l)}$ is the weight matrix at layer $l$. After quantization, this becomes:
\begin{equation}
\hat{H}^{(l+1)} = \sigma( A \hat{H}^{(l)} \hat{W}^{(l)} ),
\end{equation}
This allows faster calculations due to the lower precision. The balance between accuracy and efficiency makes low-bit quantization a viable option for faster inference.

\section{Evaluation}
\subsection{Experimental Setup}
In evaluating our framework, we utilize an AMD EPYC 7763 64-Core Processor and Ubuntu 22.04.5 LTS system. We conduct experiments on NVIDIA RTX A6000 graphics card equipped with 48 GB of GDDR6 memory, and CUDA version 12.6.

\begin{table}[h]
\centering
\caption{Hyperparameters Used in GNN Models Training}
\begin{tabular}{|l|c|}
\hline
\textbf{Hyperparameter} & \textbf{Value} \\ \hline
Learning Rate          & 0.001          \\ \hline
Epochs                 & 200            \\ \hline
Hidden Units           & 200            \\ \hline
Dropout Rate           & 0.5            \\ \hline
Batch Size             & 4              \\ \hline
Pooling Ratio          & 0.8            \\ \hline
Embedding Dimension    & 2              \\ \hline
\end{tabular}
\label{tab:hyperparameters}
\end{table}

Table~\ref{tab:hyperparameters} presents the hyperparameters used in training our GNN models. We set the learning rate to 0.001 and trained the model for 200 epochs with 200 hidden units and a dropout rate of 0.5 to prevent overfitting. A batch size of 4 was used, and we employed a pooling ratio of 0.8. Additionally, we evaluated the model every 10 epochs and used an embedding dimension of 2.

\subsection{Dataset}

\begin{figure}[htbp]
\centering
\includegraphics[width=0.8\columnwidth]{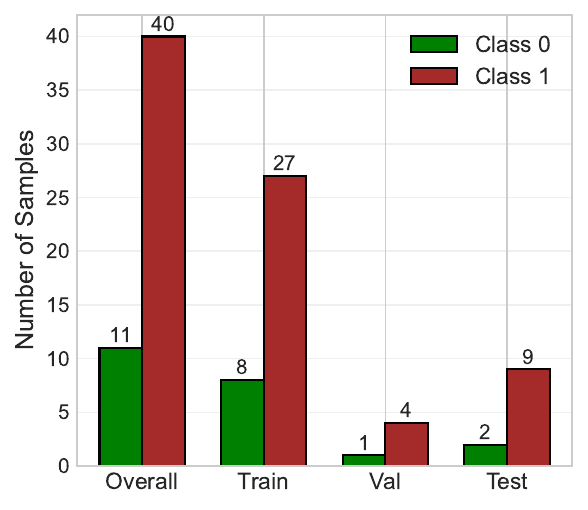}
\caption{Dataset distribution}
\label{fig:datset_ditribution}
\end{figure}

We create a dataset from the Trusthub benchmark \cite{TrustHubTrojan} and the open-source GitHub repository \cite{darkriscv}. The dataset consists of 51 different RTL designs with inserted trojans that cover denial of service, information leakage, functionality change, and performance degradation.
As shown in Figure~\ref{fig:datset_ditribution}, the dataset includes 40 designs with trojans (class 1) and 11 designs without trojans (class 0). We split the dataset into 70\% for training, 10\% for validation, and 20\% for testing the GNN model. The training set consists of 27 class 1 designs and 8 class 0 designs. The validation set includes 5 class 1 designs and 1 class 0 design. The testing set contains 9 class 1 designs and 2 class 0 designs. This distribution ensures that the dataset is diverse and balanced for training and testing the GNN models.

\begin{table}[h!]
\centering
\caption{Graph Stats for Selected Designs}
\begin{tabular}{@{}lllll@{}}
\toprule
Design          & Type   & Nodes & Edges & Time (s) \\ \midrule
AES-T1900       & TjIn   & 20446 & 25036 & 24.5754  \\
Darkriscv-T200  & TjFree & 1587  & 2175  & 1.5074   \\
Darkriscv-T100  & TjIn   & 1538  & 2098  & 1.2373   \\
AES-T1500       & TjIn   & 21100 & 25845 & 24.9904  \\
PIC16F84-T300   & TjIn   & 2636  & 3694  & 3.2588   \\ 
\bottomrule
\end{tabular}
\label{tab:design_stats}
\end{table}

\subsection{Framework evaluation}

\begin{figure*}[ht]
    \centering
    \includegraphics[width=0.99\textwidth]{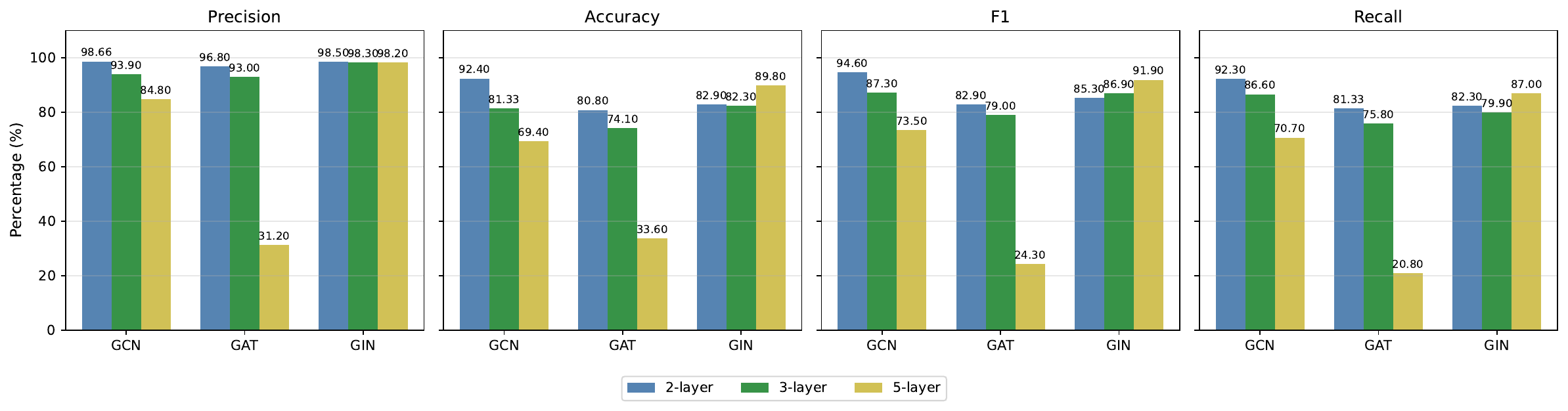}
    \caption{Evaluation results in comparison with different models: (a) Accuracy, (b) Precision , (c) F1-Scores, (d) Recall }
    \label{fig:results}
\end{figure*}

We evaluate our framework using a diverse dataset to demonstrate its effectiveness in detecting hardware trojans. To address the class imbalance, where class 0 (designs without trojans) is underrepresented, we employ a 5-fold cross-validation method in our experiments. By averaging the results over five runs, we follow standard practices to prevent overfitting and ensure that our reported metrics accurately reflect the model's performance.
We tested various GNN architectures with different layer configurations, including 2-layer, 3-layer, and 5-layer models for GCN, GAT, and GIN. Figure~\ref{fig:results} shows the evaluation results, where the x-axis represents the different GNN types and their respective layers, and the y-axis shows the percentage values of each performance metric.

 Figure~\ref{fig:results} (a) shows the precision which measures the proportion of correctly identified trojans out of all instances flagged by the model.  the 2-layer GCN model achieves the highest precision of 98.66\%, indicating it is highly effective in minimizing false positives. Additionally, the 2-layer GIN model also demonstrates high precision at 98.5\%.  As the number of layers increases, precision decreases in each model. For example, GCN precision drops from 98.66\% with 2 layers (GCN-2) to 93.9\% with 3 layers (GCN-3), and further to 84.8\% with 5 layers (GCN-5). Similarly, GAT and GIN models show a decline in precision with deeper layers, highlighting that deeper networks tend to have lower precision due to increased complexity and potential overfitting.
Accuracy measures the overall correctness of the model by calculating the proportion of true results (both true positives and true negatives) among the total number of cases examined. Figure~\ref{fig:results} (b) shows that the 2-layer GCN model also achieves the highest accuracy of 92.4\%. The 5-layer GIN model follows with an accuracy of 89.8\%. As the number of layers increases, accuracy decreases across GCN and GAT types. For instance, GCN-3 and GCN-5 achieve accuracies of 81.33\% and 69.4\%, respectively. The GIN-5 shows 89.8\% which due to potential overfitting. This decline indicates that deeper GNN models may struggle to maintain performance as model complexity increases. Figure~\ref{fig:results} (c) shows F1-score is the harmonic mean of precision and recall, providing a balance between the two metrics.  The 2-layer GCN model achieves the highest F1-score of 94.6\%, indicating a strong balance between precision and recall. The 5-layer GIN model also achieves a notable F1-score of 91.9\%. As the number of layers increases, the F1-score decreases for all GNN types. For example, GCN-3 and GCN-5 have F1-scores of 87.3\% and 73.5\%, respectively. This trend highlights that deeper models may compromise the balance between correctly identifying trojans and minimizing false positives. Figure~\ref{fig:results} (d) shows Recall, also known as the true positive rate, measures the proportion of actual trojans correctly identified by the model. The 2-layer GCN model achieves the highest recall of 92.30\%, ensuring that most trojans are detected. The 5-layer GIN model also shows a strong recall of 87.0\%. Increasing the number of layers leads to a decrease in recall, with GCN-3 and GCN-5 showing recall rates of 86.6\% and 70.7\%, respectively. Similar decreases are observed in GAT and GIN models, indicating that deeper layers reduce the model's ability to identify all true trojans.

Considering all metrics accuracy, precision, recall, and F1-score, GCN-2 consistently outperforms the other models. GCN-2 achieves the highest precision (98.66\%), accuracy (92.4\%), recall (92.30\%), and F1-score (94.6\%), demonstrating its superior capability in effectively detecting hardware trojans while maintaining a low rate of false positives. Additionally, the 5-layer GIN model shows strong performance in accuracy and F1-score, making it a competitive alternative. The decline in performance with increased layers across all GNN types suggests that shallower models are more suitable for this task, likely due to better generalization and reduced complexity in training.

\subsection{ Memory Utilization}
\begin{table}[h]
\centering
\caption{GPU Memory Usage by Different GNN Models}
\begin{tabular}{|l|c|c|}
\hline
\textbf{GNN Type} & \textbf{Layers} & \textbf{GPU Memory (MiB)} \\ \hline
GCN               & 2               & 1,688                      \\ \hline
GCN               & 3               & 1,880                      \\ \hline
GCN               & 5               & 2,322                      \\ \hline
GIN               & 2               & 2,742                      \\ \hline
GIN               & 3               & 3,472                      \\ \hline
GIN               & 5               & 4,940                      \\ \hline
GAT               & 2               & 3,310                      \\ \hline
GAT               & 3               & 4,000                      \\ \hline
GAT               & 5               & 5,296                      \\ \hline
\end{tabular}
\label{tab:gpu_memory}
\end{table}

Table~\ref{tab:gpu_memory} presents the GPU memory usage for different GNN models with varying layer configurations. The results show that GPU memory consumption increases with the number of layers for all GNN types. Specifically, the GAT model has the highest memory usage, starting at 3,310 MiB for 2 layers and escalating to 4,000 MiB with 3 layers and 5,296 MiB with 5 layers. the GCN model uses 1,688 MiB with 2 layers, which increases to 1,880 MiB with 3 layers and 2,322 MiB with 5 layers. The GIN model starts at 2,742 MiB for 2 layers and rises to 3,472 MiB for 3 layers and 4,940 MiB for 5 layers. These findings indicate that deeper GNN architectures require significantly more GPU memory, which is an important consideration for scalability and deployment in environments with limited resources. In subsection Quantized Model Results \ref{quant}, we discuss memory saving.
\subsection{Comparison with SOTAs}

\begin{figure*}[ht]
    \centering
    \includegraphics[width=0.98\textwidth]{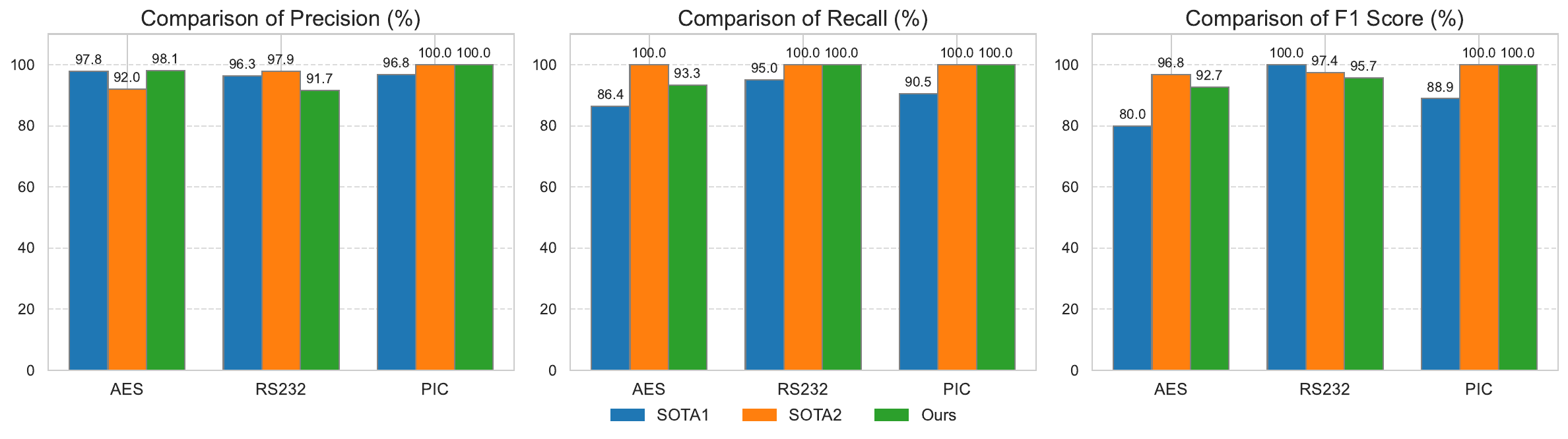}
    \caption{Evaluation results in comparison with SOTAs: (a) Precision, (b) Recall, (c) F1-Scores; SOTA1=\cite{yasaei2022hardware} and SOTA2=\cite{10560302}}
    \label{fig:comparision}
\end{figure*}

To demonstrate our framework's performance, we compared it with two GNN-based state-of-the-art (SOTA) methods: SOTA1 \cite{yasaei2022hardware} and SOTA2 \cite{10560302}. For a fair comparison, we set the same hyperparameters (learning rate: 0.001, epochs: 200, hidden units: 200) and used the same GNN layers (i.e., a 2-layer GNN model). In these comparison experiments, we did not employ the validation dataset split and k-fold validation technique for fair comparison. This might lead to overfitted models and we tried our best to provide non-overfit results for a fair comparison with SOTAs. As shown in Fig.~\ref{fig:comparision}, we present the metrics of interest on the y-axis and the designs (AES, RS232, and PIC) on the x-axis. Note that these papers only used a subset of our dataset, so we created the same dataset as the SOTAs to show the results.
We show precision comparison in Fig.~\ref{fig:comparision}~(a) with the SOTAs. Our framework outperforms the SOTAs with 98.1\% precision for the AES and  for RS232 is slightly lower, which can be attributed to fewer training samples for the RS232 designs in our dataset. The PIC dataset is too small models give the overfitted for multiple ran result as evidenced by SOTA 2 and our results.
Fig.~\ref{fig:comparision}~(b) shows recall comparison. As we can see, our framework achieves the best recall 93.3\% for AES designs. For other designs model is overfitting the results. Fig.~\ref{fig:comparision}~(c) shows F1 scores, where our scores are better than SOTA1  for AES and RS232 dataset which non a overfitted model. Overall, our framework outperforms the SOTA methods in most of the evaluation metrics.

\subsection{Quantized Model Results}
\label{quant}
\begin{figure}[htbp]
\centering
\includegraphics[width=\linewidth]{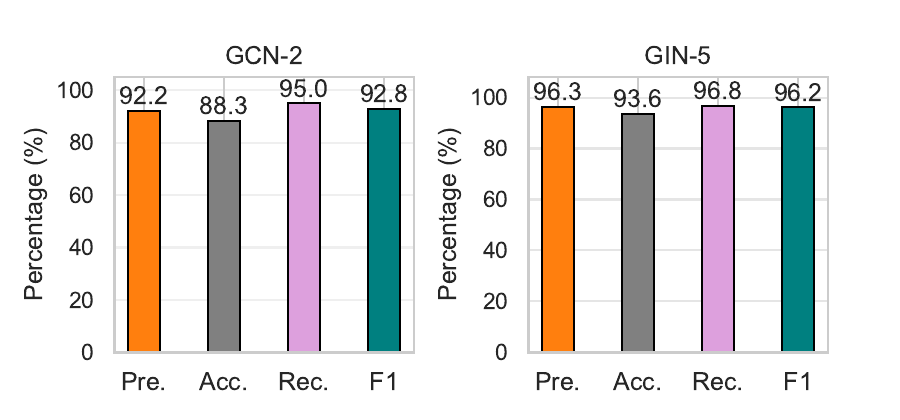}
\caption{Quantized GNN model results with: (a) GCN-2 , (b) GIN-5 }
\label{fig:quantization}
\end{figure}

To enhance the efficiency of our framework for training and inference, we quantize our model. The quantized model uses significantly less GPU memory to process the GNN. We apply 4-bit quantization using the \texttt{bitsandbytes} library \cite{dettmers2023spqr}, which reduces our 32-bit model weights to 4-bit precision while maintaining similar accuracy. This process decreases the memory footprint, speeds up inference, and facilitates deployment on hardware with limited resources. In 4-bit quantization, weights are limited to $2^4 = 16$ levels, reducing storage needs by a factor of 8 compared to 32-bit floating-point weights \cite{Jacob2017QuantizationAT, Dettmers2023QLoRAEF}.
To demonstrate the framework's performance with quantization, we selected the two best-performing models from Figure~\ref{fig:results}: the 2-layer GCN and the 5-layer GIN models. Figure~\ref{fig:quantization} (a) shows the quantized 2-layer GCN model. In the quantized version, precision is 92.2\%, compared to 98.66\% in the non-quantized model. Similarly, accuracy decreases from 92.4\% to 88.3\%. These slight reductions in performance are outweighed by the substantial decrease in GPU memory usage.
Figure~\ref{fig:quantization} (b) shows the quantized 5-layer GIN model, which uses 4,940 MiB of GPU memory as shown in Table~\ref{tab:gpu_memory}. In the quantized version, precision is 96.3\%, compared to 98.2\% in the non-quantized model. Although there is a minor decrease in precision, the quantized model benefits from significantly reduced GPU memory consumption.
These results indicate that quantization slightly lowers some performance metrics but offers substantial benefits in terms of memory efficiency. This trade-off makes our framework suitable for scaling to larger graphs with complex features, enabling deployment in environments with limited GPU resources without a significant loss in detection performance.

\section{Conclusion}

In this paper, we developed a novel framework for detecting hardware trojans in large-scale chip designs, such as RISC-V cores. Our framework includes graph extraction to enhance coverage and explores various GNN models to identify the most effective ones for HT detection. To improve training and inference efficiency, we applied 4-bit quantization to the models, significantly reducing GPU memory usage while maintaining high accuracy. We evaluated our framework using a custom dataset, achieving a precision of 98.66\% and a recall of 92.30\%. These results demonstrate the effectiveness and efficiency of our approach in accurately identifying hardware trojans in complex chip designs, making it suitable for deployment in resource-constrained environments.

\bibliographystyle{plain}
\bibliography{Sections/ref}

\end{document}